\documentclass[10pt,twocolumn,letterpaper]{article}

\usepackage[pagenumbers]{cvpr} 

\definecolor{cvprblue}{rgb}{0.21,0.49,0.74}
\usepackage[pagebackref,breaklinks,colorlinks,allcolors=cvprblue]{hyperref}
\usepackage{multirow}
\usepackage{multicol}
\usepackage{subcaption}
\usepackage[table]{xcolor}
\usepackage{cuted}

\def\paperID{*} 
\def\confName{CVPR}
\def\confYear{2026}

\title{IIR-VLM: In-Context Instance-level Recognition \\for Large Vision-Language Models}

\author{
Liang Shi$^{1}$ \quad Wei Li$^{2}$ \quad Kevin M Beussman$^{2}$ \quad Lin Chen$^{2}$ \quad Yun Fu$^{1}$\\
$^{1}$Northeastern University \quad $^{2}$Wyze Labs, Inc.\\
{\tt\small shi.lia@northeastern.edu, \{wei.li, kevin.beussman, lchen\}@wyze.com, yunfu@ece.neu.edu}
}

\begin{document}
\maketitle
\begin{abstract}
Instance-level recognition (ILR) concerns distinguishing individual instances from one another, with person re-identification as a prominent example. Despite the impressive visual perception capabilities of modern VLMs, we find their performance on ILR unsatisfactory, often dramatically underperforming domain-specific ILR models. This limitation hinders many practical application of VLMs, e.g. where recognizing familiar people and objects is crucial for effective visual understanding. Existing solutions typically learn to recognize instances one at a time using instance-specific datasets, which not only incur substantial data collection and training costs but also struggle with fine-grained discrimination. In this work, we propose IIR-VLM, a VLM enhanced for In-context Instance-level Recognition. We integrate pre-trained ILR expert models as auxiliary visual encoders to provide specialized features for learning diverse instances, which enables VLMs to learn new instances in-context in a one-shot manner. Further, IIR-VLM leverages this knowledge for instance-aware visual understanding. We validate IIR-VLM's efficacy on existing instance personalization benchmarks. Finally, we demonstrate its superior ILR performance on a challenging new benchmark, which assesses ILR capabilities across varying difficulty and diverse categories, with person, face, pet and general objects as the instances at task.
\end{abstract}

\section{Introduction}
\label{sec:intro}

Instance-level Recognition (ILR)~\cite{ilr-met, ilr-landmark, ilr-google-landmark, ilias, uned, sop, personreid} requires models to distinguish specific instances rather than general categories. A prominent example is person re-identification~\cite{personreid, transreid, chatreid}, which focuses on identifying the same person across images from different cameras or in new environments. This challenge of fine-grained matching naturally extends to other categories, such as recognizing specific pets~\cite{petface}, vehicles~\cite{transreid}, faces~\cite{vggface}, or general objects~\cite{ilias, uned, sop}. Together, these tasks illustrate a fundamental demand of visual perception, which is to differentiate specific instances across diverse domains. 

Despite the impressive visual understanding capabilities of large vision-language models (VLMs), we find that they struggle with instance-level recognition. When prompted to identify the same instance from a small gallery of images, even state-of-the-art VLMs often fail to select the correct instance, especially when the gallery images exhibit high visual similarity. A wide range of real-world applications can be affected by such an inability. For example, users of a smart-home camera may expect their VLM to reliably distinguish between family members, or tell their own pet apart from their neighbor's similar-looking pet. Instance-level recognition capabilities must serve as a cornerstone of such personalization features. 

To enable the identification of instances, most existing solutions leverage specialized, per-instance training~\cite{myvlm, yollava, mcllava}. While effective, this paradigm is relatively inefficient, requiring the curation of per-instance datasets and a repetitive training cycle for every new instance. Beyond per-instance training, specialized methods like IDA-VLM~\cite{idavlm} explore instance understanding without repetitive fine-tuning cycles. However, their demonstrated improvement is largely confined to face datasets, showing limited success on the diverse and challenging task of differentiating instances with similar visual appearances, which our work seeks to improve.


We overcome the limitations of existing work by creating In-context Instance-level Recognition VLM, or IIR-VLM, which is required to to learn and recognize new instances on-the-fly. Our framework achieves this with two primary components. First, we elevate the VLM's perception to fine-grained, instance-discriminating details by integrating instance-level recognition expert models as auxiliary visual encoders. This component addresses the fact that the VLM's original visual encoder, while strong on general semantics, is not trained to distinguish fine-grained instances and thus lacks the discriminative power for highly similar instances. In light of this, we introduce an auxiliary encoder mechanism. Through an attention-based adapter, we fuse the highly instance-discriminative features from an expert (e.g., an off-the-shelf person re-identification model) with those from the VLM's original, general-purpose visual encoder. This process transfers the experts' ability to identify subtle distinguishing clues, allowing the VLM's performance to become comparable to, or even surpass, that of the experts themselves. We quantitatively demonstrate the auxiliary encoder's critical role on fine-grained tasks, showing its performance advantage over the baseline widens as the tasks involve more visually similar instances. 

Second, we introduce a two-stage, lightweight training process to adapt the VLM to challenging ILR data. We repurpose existing ILR datasets of diverse categories into an instruction-tuning format, where each task consists of a query and a small image gallery, essentially asking the model to learn instances from the gallery in-context. This process provides a comprehensive training dataset that spans varying difficulty and instance categories. In Stage 1, the model is trained simply to identify the matching instance from the gallery. Stage 2 then advances this capability by training the model to not only match the instance, but also leverage this new instance-level knowledge to perform related visual understanding tasks. This training paradigm empowers the VLM to identify and reason about any new instance of different categories provided in-context, eliminating the need for the costly, per-instance fine-tuning of previous methods.

Extensive experiments validate our IIR-VLM's capabilities, showing its clear advantages over existing methods on instance learning benchmarks~\cite{idavlm, myvlm, yollava}. We further evaluate our method on our benchmark of challenging, fine-grained ILR tasks spanning multiple difficulty levels and categories, including persons, faces, pets, and general objects. Strong performance on these tasks promises our model's effectiveness for real-world applications, such as smart-home cameras, where distinguishing specific instances is essential for all-around personalized visual understanding.

We summarize our contribution as follows:

(1) We propose a simple and effective method to integrate instance-level recognition models as auxiliary visual encoders to VLMs, infusing the VLM with powerful instance-discriminating capabilities. 

(2) Based on the new architecture, we train IIR-VLM to perform In-context Instance-level Recognition through a lightweight, two-stage fine-tuning process, allowing VLMs to reliably learn and identify new instances on-the-fly. 

(3) We establish a comprehensive instance-level recognition (ILR) benchmark for VLMs, spanning varying difficulty levels and categories. On the benchmark, our IIR-VLM outperforms strong baselines and prior specialized methods in both instance matching and instance-aware visual understanding.

\section{Related Work}
\label{sec:related_work}

\textbf{Instance-level recognition and retrieval. } Instance-level recognition (ILR) tasks~\cite{ilr-google-landmark, ilr-met, ilr-landmark, ilias, uned} require models to identify specific instances from visually similar candidates. The most prominent example is person re-identification (Re-ID)~\cite{personreid, transreid, market, prcc, rstp}, which focuses on matching individuals across different views. In terms of method, most work leverage comparison of instance embeddings, which are obtained by training models on instance classification, and are often jointly optimized with metric learning objectives to enforce separation between features of different instances. This core challenge of Re-ID has been extended to other specific domains, such as pet Re-ID~\cite{petface} and vehicle Re-ID~\cite{transreid}. Similarly, instance-level retrieval tasks are developed for domains like landmarks~\cite{ilr-landmark}, artwork~\cite{ilr-met}, and online products~\cite{sop}, which all share the common goal of fine-grained instance matching.

More recently, the field has explored non-category-specific ILR~\cite{ilias, uned}. These methods aim to generate robust identity embeddings for any object, not just those from a predefined category. UnED~\cite{uned} proposes to learn a single embedding model for diverse instance retrieval tasks. ILIAS~\cite{ilias} builds a challenging benchmark for general instance retrieval, and verifies that pre-training on ILR tasks significantly improves the performance of general-purpose visual encoders. This finding directly motivates our work, as we show effectiveness of an auxiliary ILR expert to supplement the VLM's general visual encoder.

\textbf{Vision-language models and visual encoders. }
Large vision-language models (VLMs)~\cite{llava, qwen2.5vl, deepseekvl} have achieved remarkable success in general visual perception, excelling at tasks such as generating detailed captions, answering open-ended questions, and performing complex visual reasoning.
In a typical VLM, a visual encoder maps features to the token space of a large language model (LLM) via a projector, serving as the model's visual perception unit. While many different general-purpose visual encoders are considered in prior work~\cite{llava, qwen2.5vl, deepseekvl}, evidence~\cite{cambrian} reveals that this choice has a selective impact on downstream performance. Specifically, an encoder's pretext task, such as vision-language alignment~\cite{clip, siglip, siglip2} or self-supervised learning~\cite{dino, dinov2}, determines its suitability for different VLM tasks. For example, VLMs based on language-supervised encoders perform far better on OCR and chart-related tasks~\cite{cambrian}. Furthermore, features from different encoders can be aggregated to complement each other and accomplish tasks where single encoders fail~\cite{eyeswideshut, cambrian}.

In this work, we argue that instance-level recognition (ILR), much like OCR, is a highly specialized task. General-purpose encoders perform poorly on it due to their lack of necessary, category-specific fine-grained features. We demonstrate that a VLM's ILR capabilities can be enhanced by complementing the general encoder with small, specialist models trained specifically for this purpose.

\textbf{VLM personalization. }In training VLMs to understand specific instances, existing work mainly relies on per-instance training~\cite{myvlm, yollava, mcllava}. This approach collects a small image set for each new instance and then fine-tunes the VLM to associate those visual features with a unique instance token, which the model uses to refer to that instance. Specifically, MyVLM~\cite{myvlm} trains individual concept heads that indicate the existence of instances, before deciding to include the instance token as part of its response. Yo'LLaVA~\cite{yollava} trains the instance token as well as a set of description tokens to help the VLM encode crucial information of the instance. MC'LLaVA~\cite{mcllava} tackles multiple instance appearances. While effective, per-instance training is considered inefficient. It usually requires not only the curation of per-instance datasets including positive and negative samples, but also additional training for every new instance. Training on single instances also leads to struggles with fine-grained ILR tasks that require differentiating between highly visually similar images. 

A few studies, primarily in face understanding, pioneer instance understanding without per-instance training. PLVM~\cite{plvm} uses additional general-purpose encoders for gallery image encoding. IDA-VLM~\cite{idavlm} applies a specifically designed encoder that involves explicit cross-image comparison for face-related movie understanding. Our work generalizes these approaches to diverse and challenging ILR tasks, identifying dedicated ILR-trained encoders as the critical component for succeeding at challenging, fine-grained instance-level recognition.

\section{Method}

\begin{figure}[t]
    \centering
    \includegraphics[width=1\columnwidth]{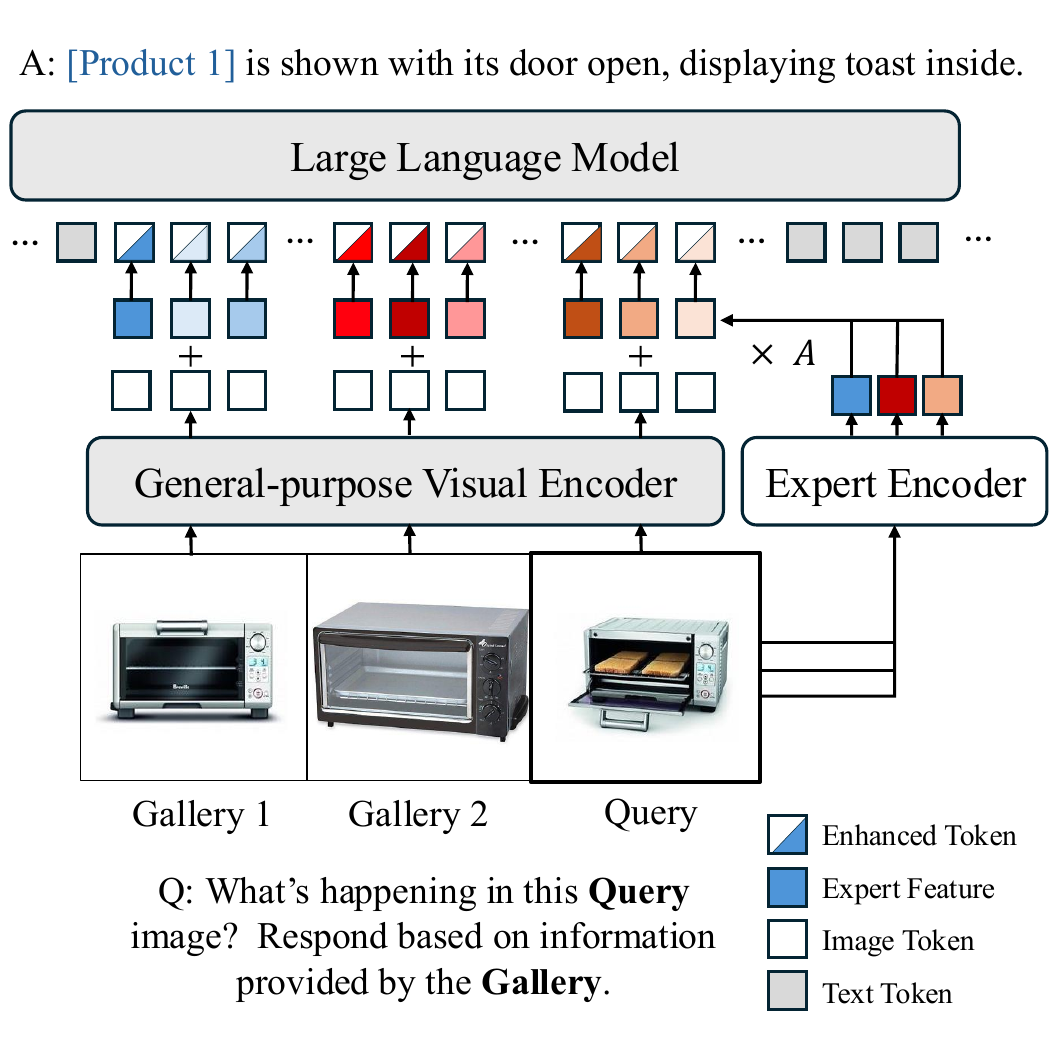}
    \caption{\textbf{Pipeline of our proposed IIR-VLM. } IIR-VLM learns instances in-context from the gallery and performs visual understanding tasks related to the query. All images are processed by both the original general-purpose visual encoder and an instance-level recognition expert encoder, which is trained exclusively on instance classification tasks. Expert features are re-weighted by an attention score $A$ to indicate instance presence in each original feature token. Re-weighted features are added to the original tokens to provide discriminative features beneficial for ILR tasks.  }
    \label{fig:pipeline}
\end{figure}

This paper proposes IIR-VLM, which advances large VLMs with an In-context Instance-level Recognition capability. We achieve this through a dual-component framework. We first integrate instance-level recognition models as auxiliary visual encoders, which provides category-specific knowledge for instance discrimination (\cref{method: expert}). This enhanced VLM is then efficiently fine-tuned in a two-stage process (\cref{method: data}): aligning the feature spaces through challenging in-context instance matching in Stage 1, and enhancing the model with instance-aware understanding tasks in Stage 2. The complete pipeline is illustrated in~\cref{fig:pipeline}.

\subsection{Preliminary}

A large Vision-Language Model (VLM) combines a visual encoder with a Large Language Model (LLM) through a projector, and works to process interleaved visual and text inputs to perform various visual perception tasks. Specifically, given a series of input text $X=(x_1, x_2, \cdots , x_{n-1})$ and image inputs $c$, VLMs predict
\begin{equation}
P(x_n|f(c), x_1, x_2, \cdots , x_{n-1}),
\end{equation}
where $f(\cdot)$ denotes visual encoders, in order to decode responses based on inputs. 

This paper investigates and advances the capabilities of VLMs on the in-context instance-level recognition (ILR) task~\cite{ilr-landmark, ilr-met, ilr-google-landmark}. ILR can be described as matching images that depict the same instances across diverse categories, including persons, faces, and arbitrary objects. In a typical formulation, a query image $c_q$ is provided along with a set of gallery images $(c_g^1, c_g^2, \cdots c_g^k)$, where the goal is to identify which gallery image $c_g^i$ shares the same "identity", \ie, involves the same instance, as $c_q$. Traditional solutions involve ILR model $f_e$ that are trained to extract instance-discriminating features, and images are matched through feature similarity: 
\begin{equation}
y = \text{argmax}_i (Sim(f_e(c_q), f_e(c_g^i)).
\end{equation}
\label{eq:similarity}

Instance-level recognition, and subsequently visual understanding based on instance knowledge, are invaluable for VLMs to master, though we show that state-of-the-art models perform poorly on the task, significantly underperforming traditional ILR models. We refer to these traditional models $f_e$ as \textit{experts}, and show how they can be utilized to efficiently enhance VLMs on the task.

\subsection{Integrating ILR experts}
\label{method: expert}

\textbf{Integration mechanism}. Our method enhances the VLM's image features by injecting instance-aware features from ILR models. For each image $c$ containing a target instance of the expert, the VLM's original, general-purpose encoder provides a multi-token feature $f(c)=[f_1(c), f_2(c), \cdots f_N(c)]$, while the expert encoder extracts a single, highly discriminative one-dimensional identity feature, $f_e(c)$. To integrate the features, $f_e(c)$ is first projected through a trainable MLP into $f'_e(c)$, and then used as a query to calculate an attention score $A$ against every token in $f(c)$:

\begin{equation}
A_i = \text{Softmax}_N \left( \text{Sim}(f_i(c), f'_e(c)) \right).
\end{equation}

Inspired by the positional encoding mechanism~\cite{attention}, which injects spatial information into Transformer inputs through additive modulation, we treat the attention-enhanced feature $f'_e(c)$ as an "identity embedding". This embedding is added back to the original token features, effectively infusing fine-grained instance identity knowledge into the VLM’s visual representation:
\begin{equation}
F_i(c) = f_i(c) + A_i \cdot f'_e(c).
\end{equation}

The original encoder's multi-token features $f(c)$ are then replaced by the instance-aware feature sequence $F(c)=[F_1(c), F_2(c), \cdots F_N(c)]$ for all subsequent VLM processing. Multiple experts can be used in parallel to enhance images containing different categories of instances.

\textbf{Selection of expert encoders. } We empirically show that the identity embedding \textit{requires} a model explicitly trained on instance discrimination, instead of general visual encoders, for the best performance. We utilize off-the-shelf models for categories that are mature in instance learning, such as person~\cite{plip, transreid} and face~\cite{arcface}. For categories such as pet and household items, where off-the-shelf experts are readily not available, we train a dedicated ILR expert to meet this demand. Following ILIAS~\cite{ilias}, we fine-tune a general-purpose visual encoder with instance classification loss and metric learning losses, specifically a triplet loss that pulls features of the same instance closer and pushes different instances away~\cite{transreid}. The models are fully-finetuned to achieve maximum performance. This step ensures that the resulting encoder possesses the instance-discriminative, fine-grained feature that effectively supplements the original visual encoder.

\subsection{Data engine and two-stage training}
\label{method: data}
\textbf{Dataset formulation. } The training of VLMs requires datasets in the format of instruction tuning. To that end, we repurpose traditional instance-level recognition datasets~\cite{personreid, prcc, cuhk, vggface, sop}, which feature a large number of distinct instances and multiple images per instance. These images are grouped and embedded into user-assistant conversations, where every conversation includes a query image ($c_q$) and a gallery of candidate images ($G=\{c_g^1, c_g^2, \ldots, c_g^K\}$) that potentially contains matches for the queried instance. At inference time, the VLM is provided with the gallery in its context, and required to understand the query image based on matching results. 

While some instances can be straightforwardly identified based on their visual appearance, instance-level recognition often requires more fine-grained attention to subtle visual cues, particularly when dealing with highly similar instances. In this work, we emphasize the capability of performing challenging fine-grained tasks, which are not immediately solvable by general visual perception. Leveraging general visual encoders, we modulate the difficulty level of the dataset by filling the gallery with strong distractors, selected based on visual similarity to the query. Specifically, we require $Sim(f(c_g^i), f(c_q))>\tau$ for some pre-determined similarity threshold $\tau$. 

\textbf{Two-stage fine-tuning with ILR awareness. }With the dataset and auxiliary expert encoders, we introduce a lightweight, two-stage fine-tuning process to embed Instance-Level Recognition (ILR) awareness into the VLM. This process minimizes computational cost and mitigates overfitting by freezing the parameters of both the LLM and the visual encoders, and training only the projectors. 

\textbf{Stage 1} of the framework aims to establish the VLM's foundational ability to recognize and distinguish instances in-context. The VLM is trained to correctly identify which image in the gallery $G$ matches the identity of the query image $c_q$ provided in the prompt, essentially answering a multiple-choice question. We aim to align the expert features with the VLM in this stage for effective knowledge transfer. \textbf{Stage 2} subsequently improves the model's capability to instance-aware understanding. The VLM is tasked with generating a detailed caption of the query image $c_q$, leveraging the identity information present in the gallery $G$. For example, in a conversation that includes Person 3 of the gallery, the model is trained to respond with specific, bracketed identity, such as \texttt{[Person 3] walks into the room wearing a blue shirt.} This mechanism allows the model to anchor its generated text to the identity learned in-context, and the bracketed token can readily be replaced by an instance's name in practical applications.
\section{Experiments}

\begin{figure*}[t]
    \centering
    \includegraphics[width=1\textwidth]{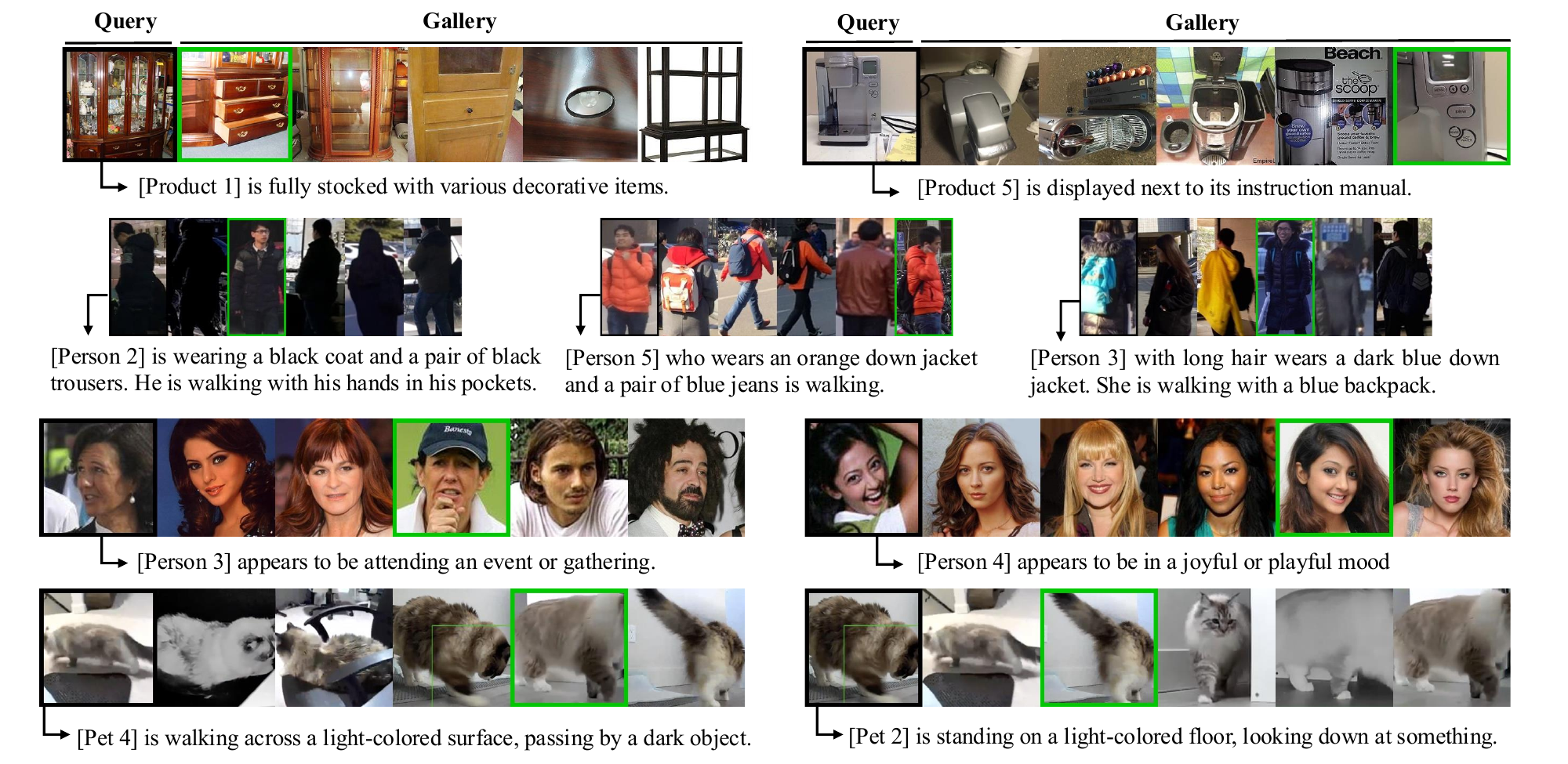}
    \caption{\textbf{Samples of our instance-level recognition benchmark and corresponding model outputs. }We cover a wide span of categories, including household items (object), persons, faces, and pets. In each conversation, models take a group of images as input, where the leftmost image represents the \textbf{query} image at task, and five visually similar \textbf{gallery} images follow. Models are required to select the gallery image with the matching instance (in green), and generate captions of the query image referring to the correct instance. }
    \label{fig:captions}
\end{figure*}

We perform a series of experiments to verify the validity of IIR-VLM to learn instances in-context. In ~\cref{exp:setup}, we outline our experimental setup, including benchmark construction, training configuration, and sources of expert encoders. We evaluate our model on our challenging ILR benchmark in~\cref{exp:our_benchmark}, compare it with existing VLM and per-instance methods in ~\cref{exp:comparisons}, and demonstrate strong results in both matching and captioning. Finally, through ablations in ~\cref{exp:ablation}, we quantify the impact of expert encoders, task difficulty, and two-stage training on overall performance.

\subsection{Setup}
\label{exp:setup}
\textbf{Datasets. } We build our instruction-tuning training and test data by re-purposing instance-level recognition datasets of different domain. We put together datasets spanning four categories, i.e. person, face, pet, and general objects. For the person category, we utilize a set of public person Re-ID dataset~\cite{market, prcc, rstp, lpw, cuhk}, which combines into 76k images. We use the VGGFace~\cite{vggface} dataset for the face category. Around 60k images are collected for the pet category. The Stanford Online Product~\cite{sop} dataset, which contains 120k images representing 23k instances across 12 household categories, is used for learning general object ILR. 

For instruction tuning, we select gallery length $K=5$ and similarity threshold $\tau=0.5$ by default to provide sufficiently challenging data. All instruction tuning datasets are split into training and test sets such that there are no overlapping identities between the two splits. The test set, featuring 500 conversations per instance category, serves as the cornerstone of our evaluation and will be released. 

Finally, we generate captions with bracketed subjects (see~\cref{method: data}) for all query images using Gemini 2.5 Flash and train models to perform captioning in the second stage. 

\textbf{Implementation details. } We use Qwen2.5-VL-3B-Instruct~\cite{qwen2.5vl} as our base model. During training, the LLM and the original visual encoders are both kept frozen. Only projectors between all encoders to the LLM are updated. Image inputs are presented as is to the VLM's original visual encoder; however, when target instance exists, images are adjusted and potentially cropped to meet input requirements for expert encoders. 

Dedicated ILR experts are selected based on category. We use off-the-shelf person re-identification models PLIP~\cite{plip} for the person category, and the Arcface~\cite{arcface} face recognition model for the face category. Experts for pet and general object are fine-tuned from general-purpose visual encoders~\cite{dinov2} on instance classification tasks.

\subsection{Challenging instance-level recognition}
\label{exp:our_benchmark}

~\cref{fig:captions} provides representative examples from our instance-level recognition benchmark along with outputs from our model. The visually challenging nature of the benchmark is highlighted by galleries containing highly similar negative samples. Many test cases are challenging even to the human eye at first glance; however, correct decisions can be made by focusing on subtle details. The model's success on these tasks is therefore determined by its ability to capture these category-specific, fine-grained details hidden in the images.

As the samples and corresponding outputs demonstrate, our VLM successfully handles this complexity, consistently identifying the correct matching instance (marked in green). Furthermore, the model validates the second stage training by producing captions for the query images (marked in black). The generated captions are consistently correct visual descriptions that cater to the instance's immediate status, context, and background, rather than reciting pre-existing knowledge of the instance. The model also correctly utilizes the bracketed subject (e.g., \texttt{[Person 3]}), a format that allows for fast replacement of real, user-specified names. In real-world applications such as smart-home visual assistants, the model can be provided in-context with a gallery of family members, important items, pets, etc., and be expected to describe captured scenes based on accurate knowledge of instances to remember.

We systematically evaluate the performance of VLMs on our four-category instance-level recognition (ILR) benchmark. We compare our IIR-VLM with state-of-the-art the VLMs Gemini 2.5 Pro, Qwen2.5-VL-3B~\cite{qwen2.5vl} (our base model), Qwen2.5-VL-7B~\cite{qwen2.5vl}, as well as IDA-VLM~\cite{idavlm}, a recent specialized VLM proposed for instance learning and movie understanding. We report the accuracy of finding the correct instance in~\cref{tab:main_table}.

\begin{table}[tbp]
  \centering
  \caption{Accuracy (\%) of four different VLMs on our benchmark, which tests models' capability on challenging instance-level recognition (ILR) tasks across four categories. }
  \label{tab:main_table}
\small
\begin{tabular}{lccccc}
\toprule
 & Object & Person & Face & Pet & Avg \\
\midrule
Gemini 2.5 Pro & 89.8 & 75.4 & 72.8 & 79.6 & 79.4 \\
Qwen2.5 VL-3B & 54.0 & 52.6 & 65.4 & 49.6 & 55.4 \\
Qwen2.5 VL-7B & 70.8 & 73.6 & 84.8 & 68.4 & 74.4 \\
IDA-VLM~\cite{idavlm} & 59.6 & 42.0 & 75.0 & 51.4 & 57.0 \\
\midrule
\rowcolor{black!5}
\multicolumn{6}{l}{\textit{Stage 1 models: trained for recognition.}} \\
Ours w/o Expert & 89.6 & 90.4 & 91.4 & 86.7 & 89.5 \\
Ours w/ Expert & \textbf{92.9} & \textbf{92.6} & \textbf{91.4} & \textbf{88.3} & \textbf{91.3} \\
\midrule
\rowcolor{black!5}
\multicolumn{6}{l}{\textit{Stage 1 + 2 models: trained for recognition and caption.}} \\
Ours w/o Expert & 89.4 & 81.6 & 85.8 & 90.9 & 86.9 \\
Ours w/ Expert & \textbf{91.2} & \textbf{84.4} & \textbf{86.2} & \textbf{92.2} & \textbf{88.5} \\
\bottomrule
\end{tabular}
\end{table}

Among the general state-of-the-art models, Gemini 2.5 Pro scores the highest average baseline performance, demonstrating particular strength in the object category, reaching 89.8\%. The open-source Qwen2.5-VL models exhibit strong performance across both model sizes. Notably, the 7B variant significantly improves accuracy, securing the second rank among the compared methods, directly following the proprietary Gemini model.

IDA-VLM~\cite{idavlm} showcases its ILR capability by performing promisingly on the face category, scoring 75.0\% and notably surpassing Gemini. However, it is evidently less capable of generalizing its ILR capability to other categories, resulting in a lower overall average. 

We separately report the performances of four variants of our VLMs, namely models trained with one or two stages, and models trained with and without experts. Training on multiple-choice questions in Stage 1 provides maximized ILR results, bringing the average accuracy to above 90\%. Training for the second stage enables a crucial ability to perform visual understanding with instance knowledge. The new capability incurs a modest performance drop of approximately 2.2\% on average, indicating an expected trade-off between maximizing ILR capabilities and general visual understanding with instance knowledge. Nevertheless, even after this trade-off, our full model with the expert encoder maintains a strong average score of 88.5\%, surpassing our base Qwen2.5-VL-3B model by 33.1\% and leading all state-of-the-art VLMs tested. 

Across both stages, our results demonstrate the necessity and value of the expert encoder. VLMs trained with ILR experts as auxiliary encoders consistently outperform their counterparts that rely only on the original encoder across all categories. Quantitatively, the expert boosts the average Stage 1 accuracy from 89.5\% to 91.3\%. By supplementing the general-purpose visual encoder with category-specific knowledge on instance discrimination, the expert encoders allow for a strong instance-discriminating capability, and establishes a new state-of-the-art for this instance-level recognition task.

\subsection{Comparison with existing solutions}
\label{exp:comparisons}

\setlength{\tabcolsep}{4pt}
\begin{table}[tbp]
  \centering
  \caption{Instance-level recognition accuracy (\%) on the matching tasks of MM-ID~\cite{idavlm}, compared with IDA-VLM~\cite{idavlm} }
  \label{tab:ida_vlm_comparison}
\small
\begin{tabular}{lccccc}
\toprule
 & Face & Animal & Building & Vehicle & Avg \\
\midrule
IDA-VLM~\cite{idavlm} & 82.1 & 58.3 & 100 & 30.0 & 74.0 \\
IIR-VLM (Ours) & \textbf{92.9} & \textbf{66.7 }& \textbf{100} & \textbf{60.0} & \textbf{85.1} \\
\bottomrule
\end{tabular}
\end{table}
\setlength{\tabcolsep}{6pt}

\textbf{Matching tasks}. In ~\cref{tab:ida_vlm_comparison}, we compare with IDA-VLM~\cite{idavlm} on their proposed MM-ID benchmark. MM-ID is a pioneering benchmark that aims to evaluate VLMs on memorizing and utilizing identity information of visual inputs, with particular focuses on face understanding in movie scenes. We test IIR-VLM on the matching task of MM-ID. Despite not having trained on movie-relevant data, our method achieves superior accuracy across all evaluated categories, demonstrating a promising generalization capability to various instance-level recognition problems.

We further compare IIR-VLM with per-instance training methods MyVLM~\cite{myvlm} and Yo'LLaVA~\cite{yollava} on their benchmark. Both benchmarks evaluate a model's ability to detect the presence of a specific, pre-trained concept within a new image, reporting accuracy on both positive (concept present) and negative (concept absent) samples. In our formulation, the task is analogous to a gallery size $K=1$ setting. We train a variant of our model which does not assume the existence of query images in the gallery to meet the demand of the task. Instead of utilizing multiple images to learn instance-specific tokens, our model learns all concepts with a single image in-context. 

As shown in \cref{tab:myvlm_yollava}, IIR-VLM achieves a weighted score of 92.8\% on Yo'LLaVA's benchmark, a result comparable to Yo'LLaVA's own score and outperforming GPT-4V. On MyVLM's benchmark, our model pushes towards the scores of MyVLM~\cite{myvlm} and Yo'LLaVA~\cite{yollava}, though a small gap remains. This gap stems largely from a lower accuracy on positive samples, where a learned instance exists in the queried image. This suggests that our model is more conservative, exhibiting a higher tendency to deny a match even when the correct instance is present. We hypothesize that this is a result of its training on our challenging ILR tasks, which has conditioned it to be highly sensitive to subtle visual differences. 

We emphasize that these comparable results are achieved without any costly per-instance training, which would require complex data curation and training procedure for each instance to learn. This strongly validates IIR-VLM, our in-context learning approach, as a more flexible and efficient paradigm. It demonstrates a clear potential to replace per-instance training, especially as models scale and are trained on more comprehensive, diverse ILR data.

\begin{table}[tbp]
  \centering
  \caption{Comparison with per-instance training methods. Results indicates the accuracy (\%) on correctly identifying the existence of specific instances in test images. }
  \label{tab:myvlm_yollava}
  \begin{tabular}{lccc}
    \toprule
    & Positive & Negative & Weighted \\
    \midrule
    \rowcolor{black!5}
    \multicolumn{4}{l}{\textit{MyVLM's benchmark.}} \\
    MyVLM~\cite{myvlm} & 96.6 & 90.9 & 93.8 \\
    Yo'LLaVA~\cite{yollava} & 97.0 & 95.7 & 96.4 \\
    IIR-VLM (Ours) & 87.4 & 90.5 & 88.8 \\
    \midrule
    \rowcolor{black!5}
    \multicolumn{4}{l}{\textit{Yo'LLaVA's benchmark.}} \\
    GPT-4V & 80.9 & 99.2 & 90.1 \\
    Yo'LLaVA~\cite{yollava} & 94.9 & 89.8 & 92.4 \\
    IIR-VLM (Ours) & 89.0 & 97.4 & 92.8 \\

    \bottomrule
  \end{tabular}
\end{table}

\begin{table}[tbp]
  \centering
  \caption{Comparison with MyVLM on instance-aware caption generation, with CLIP-Image measuring text-image alignment and CLIP-Text measuring similarity to ground truth captions. }
  \label{tab:captions}
  \small
  \begin{tabular}{lccc}
    \toprule
    & Examples & CLIP-Image & CLIP-Text \\
    \midrule
    MyVLM & 1 & 24.20 & 57.37 \\
    MyVLM & 2 & 24.91 & 61.01 \\
    MyVLM & 4 & 25.42 & 62.61 \\
    \midrule
    IIR-VLM (Ours) & 1 & \textbf{27.06} & \textbf{69.81} \\
    \bottomrule
  \end{tabular}
\end{table}

\textbf{Captioning tasks}. We showcase the effectiveness of our Stage 2-trained models in producing accurate captions for matched instances, in comparison with existing work. Results are shown in~\cref{tab:captions}. Following MyVLM~\cite{myvlm}, generated captions are evaluated using CLIPScore~\cite{clipscore}, which measures the alignment of the generated caption and the query image (CLIP-Image). We also evaluate the caption's fidelity to the ground truth using the CLIP text encoder (CLIP-Text). Despite learning instances in-context with only a single example image, our generated captions score higher than MyVLM on both metrics, signaling accurate description of the scene. When combined with successful instance-level recognition, the model is fully capable of identifying the correct subject and performing identity-aware visual understanding tasks.

\subsection{Ablation study}
\label{exp:ablation}

We study the effect of both critical components of our framework, namely introducing ILR expert as auxiliary encoders and two-stage training. We strengthen the argument of applying experts by testing the framework on varying difficulty levels of our ILR benchmark. 

\begin{table}
    \centering
    \caption{Accuracy (\%) on instance-level recognition tasks of visual encoders and VLMs equipped with different encoders.}
    \label{tab:results_ilr}
    \begin{tabular}{l c cc} 
        \toprule
        Model & Auxiliary Encoder & Stage 1 & Stage 1+2 \\ 
        \midrule
        \rowcolor{black!5}
        \multicolumn{4}{l}{\textit{ILR with Embedding Similarity}} \\
        DINOv2 & -- & 58.5 & -- \\ 
        Expert & -- & 91.4 & -- \\ 
        \midrule
        \rowcolor{black!5}
        \multicolumn{4}{l}{\textit{ILR with VLM.}} \\
        VLM  & None & 89.6 & 89.4 \\ 
        VLM & DINOv2 & 91.1 & 90.2 \\ 
        VLM & Expert & \textbf{92.9} & \textbf{91.2} \\ 
        \bottomrule
    \end{tabular}
\end{table}

\textbf{VLM with expert encoders.} We support the necessity of our dual-encoder design with ablation study in~\cref{tab:results_ilr}. We study the performance on object~\cite{sop} ILR tasks of standalone visual encoders and VLMs with varying visual encoder design. For a simple visual encoder like DINOv2~\cite{dinov2}, we resort to traditional Re-ID protocols. We apply it for instance-level recognition by extracting features and locating the gallery image with the highest similarity, as described in~\cref{eq:similarity}. 

In order to have an ILR expert on general object, we fine-tune a general-purpose visual encoder, DINOv2~\cite{dinov2}, with instance classification and metric learning losses. We show that the ILR-trained expert dramatically outperforms the general-purpose DINO encoder, improving from 58.5\% to above 90\%. This confirms that while general-purpose visual encoders extract rich semantics from visual inputs, they are insufficient for fine-grained ILR tasks. We study how the features of these different encoders impact the VLM's performance as auxiliary encoders. 

We evaluate VLMs on the same benchmark. When fusing an auxiliary encoder, we note that adding DINO provides a marginal improvement over the original encoder alone. This suggests that while feature fusion is beneficial by providing supplementary information, DINO's general features are not the optimal choice.

Our proposed IIR-VLM, \ie, VLM with expert model achieves the highest performance at 92.9\%, outperforming both the original fine-tuned VLM and the DINO-supported VLM. This result clearly demonstrates the necessity of using a specialized ILR-trained expert to achieve maximum performance in this challenging, fine-grained task. These performance trends are consistent across both training stages, as the advantage of experts is maintained when the ILR task is generalized to general visual understanding tasks with instance knowledge. We also note that fusing the expert model into IIR-VLM leads to an instance matching accuracy 1.5\% higher than the expert model itself. This result reveals a promising synergy between encoders, whose complementary features are effectively fused by our model to achieve performance better than any single encoders. 

\textbf{Expert advantage on challenging tasks. } Our similarity threshold parameter $\tau$ determines the difficulty of ILR tasks. We investigate how this parameter affects model performance. VLMs are trained and tested on datasets with $\tau=0.2$, $\tau=0.5$, and $\tau=0.8$. As the threshold grows, gallery images are increasingly similar to the query image, thus providing more challenges to ILR. 

\begin{figure}[t]
    \centering
    \includegraphics[width=1\columnwidth]{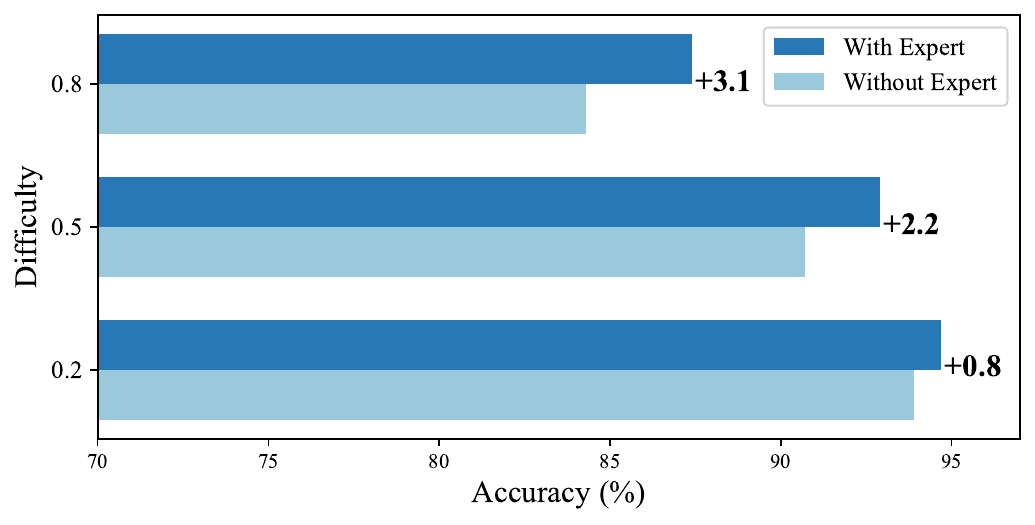}
    \caption{Comparison between models with and without ILR expert encoder across different difficulty levels. 
    Numbers on the right indicate the relative accuracy improvements.}
    \label{fig:expert_improvement}
\end{figure}

With the increase of difficulty, we show a widening advantage of VLMs trained with the auxiliary expert input, as shown in~\cref{fig:expert_improvement}. At the lowest difficulty ($\tau=0.2$), the expert provides a modest gain of +0.8\%. As the task becomes more challenging ($\tau=0.5$), the performance advantage significantly widens to +2.2\%. The trend continues at the highest difficulty level ($\tau=0.8$), where the expert provides its maximal benefit at +3.1\%. This quantitatively validates that the features provided by the expert become increasingly critical as visual similarity rises. Unlike features provided by general-purpose visual encoders, the expert features possess the precise discriminative power required to isolate subtle differences within groups of visually similar instances.

\begin{table}[t]
    \centering
    \caption{Ablation study on the effect of training stages. Results are reported in accuracy (\%).}
    \label{tab:stage_ablation}
    \begin{tabular}{lcc}
        \toprule
        \textbf{Training stages} & \textbf{w/o expert} & \textbf{w/ expert} \\
        \midrule
        Stage 2 only & 81.8 & 88.2 \\
        Stage 1 + 2 & \textbf{89.4} & \textbf{91.2} \\
        \bottomrule
    \end{tabular}
\end{table}

\textbf{Two-stage training.} We analyze the impact of our two-stage training paradigm. As shown in \cref{tab:stage_ablation}, while it is possible to directly train models to output instance-specifying captions (Stage 2 only), our results demonstrate this approach is suboptimal, yielding scores of 81.8\% (without expert) and 88.2\% (with expert). Clearly, training with expert here provides even more significant advantage. 

We find that first training on the fundamental capability of the ILR matching task, in our case the Stage 1 multiple-choice questions, is largely helpful for the final generative task. This initial matching stage boosts performance by a significant 7.6\% for the model without the expert, and 3.0\% for the model with the expert. This confirms that grounding the model in challenging instance matching tasks is a critical step for achieving robust, high-performance instance-aware visual understanding.

\section{Conclusion}

In this work, we propose IIR-VLM, a novel VLM that successfully performs In-context Instance-level Recognition, learning instances without the need for costly per-instance training. We achieve this through a modular framework that integrates pre-trained instance-level recognition expert encoders with a lightweight, two-stage training paradigm. We validate our approach on a new, challenging instance-level recognition benchmark spanning diverse categories and difficulty levels, where IIR-VLM outperforms strong general-purpose and specialized baselines. A key finding through the process is a widening advantage as the task difficulty increases, highlighting the expert as a crucial component for discriminating visually similar instances.

Our model's success on matching tasks serves as a strong foundation for future work. A crucial next step includes expanding our model's capabilities to instance-aware visual grounding, training it to not only learn and recognize an instance in-context but also be able to spatially localize it within an image or video. Successful grounding would unlock a richer set of real-world applications demanding instance understanding capabilities.
{
    \small
    \bibliographystyle{ieeenat_fullname}
    \bibliography{main}

@String(CVPR= {IEEE Conf. Comput. Vis. Pattern Recog.})

@String(ICCV= {Int. Conf. Comput. Vis.})

@String(ICPR = {Int. Conf. Pattern Recog.})

@String(AAAI = {AAAI})

@String(CVPR  = {CVPR})

@String(ICCV  = {ICCV})

@String(ICPR  = {ICPR})

@inproceedings{myvlm,
  title={Myvlm: Personalizing vlms for user-specific queries},
  author={Alaluf, Yuval and Richardson, Elad and Tulyakov, Sergey and Aberman, Kfir and Cohen-Or, Daniel},
  booktitle={European Conference on Computer Vision},
  pages={73--91},
  year={2024},
  organization={Springer}
}

@article{yollava,
  title={Yo'llava: Your personalized language and vision assistant},
  author={Nguyen, Thao and Liu, Haotian and Li, Yuheng and Cai, Mu and Ojha, Utkarsh and Lee, Yong Jae},
  journal={Advances in Neural Information Processing Systems},
  volume={37},
  pages={40913--40951},
  year={2024}
}

@article{mcllava,
  title={Mc-llava: Multi-concept personalized vision-language model},
  author={An, Ruichuan and Yang, Sihan and Lu, Ming and Zhang, Renrui and Zeng, Kai and Luo, Yulin and Cao, Jiajun and Liang, Hao and Chen, Ying and She, Qi and others},
  journal={arXiv preprint arXiv:2411.11706},
  year={2024}
}

@inproceedings{plvm,
  title={PLVM: A tuning-free approach for Personalized Large Vision-Language Model},
  author={Pham, Chau and Phan, Hoang and Doermann, David and Tian, Yunjie},
  booktitle={Proceedings of the Computer Vision and Pattern Recognition Conference},
  pages={3632--3641},
  year={2025}
}

@article{llava,
  title={Visual instruction tuning},
  author={Liu, Haotian and Li, Chunyuan and Wu, Qingyang and Lee, Yong Jae},
  journal={Advances in neural information processing systems},
  volume={36},
  pages={34892--34916},
  year={2023}
}

@article{cambrian,
  title={Cambrian-1: A fully open, vision-centric exploration of multimodal llms},
  author={Tong, Peter and Brown, Ellis and Wu, Penghao and Woo, Sanghyun and IYER, Adithya Jairam Vedagiri and Akula, Sai Charitha and Yang, Shusheng and Yang, Jihan and Middepogu, Manoj and Wang, Ziteng and others},
  journal={Advances in Neural Information Processing Systems},
  volume={37},
  pages={87310--87356},
  year={2024}
}

@inproceedings{eyeswideshut,
  title={Eyes wide shut? exploring the visual shortcomings of multimodal llms},
  author={Tong, Shengbang and Liu, Zhuang and Zhai, Yuexiang and Ma, Yi and LeCun, Yann and Xie, Saining},
  booktitle={Proceedings of the IEEE/CVF Conference on Computer Vision and Pattern Recognition},
  pages={9568--9578},
  year={2024}
}

@article{qwen2.5vl,
  title={Qwen2. 5-vl technical report},
  author={Bai, Shuai and Chen, Keqin and Liu, Xuejing and Wang, Jialin and Ge, Wenbin and Song, Sibo and Dang, Kai and Wang, Peng and Wang, Shijie and Tang, Jun and others},
  journal={arXiv preprint arXiv:2502.13923},
  year={2025}
}

@inproceedings{ilias,
  title={Ilias: Instance-level image retrieval at scale},
  author={Kordopatis-Zilos, Giorgos and Stojni{\'c}, Vladan and Manko, Anna and Suma, Pavel and Ypsilantis, Nikolaos-Antonios and Efthymiadis, Nikos and Laskar, Zakaria and Matas, Jiri and Chum, Ondrej and Tolias, Giorgos},
  booktitle={Proceedings of the Computer Vision and Pattern Recognition Conference},
  pages={14777--14787},
  year={2025}
}

@inproceedings{uned,
  title={Towards universal image embeddings: A large-scale dataset and challenge for generic image representations},
  author={Ypsilantis, Nikolaos-Antonios and Chen, Kaifeng and Cao, Bingyi and Lipovsk{\`y}, M{\'a}rio and Dogan-Sch{\"o}nberger, Pelin and Makosa, Grzegorz and Bluntschli, Boris and Seyedhosseini, Mojtaba and Chum, Ond{\v{r}}ej and Araujo, Andr{\'e}},
  booktitle={Proceedings of the ieee/cvf international conference on computer vision},
  pages={11290--11301},
  year={2023}
}

@inproceedings{transreid,
  title={Transreid: Transformer-based object re-identification},
  author={He, Shuting and Luo, Hao and Wang, Pichao and Wang, Fan and Li, Hao and Jiang, Wei},
  booktitle={Proceedings of the IEEE/CVF international conference on computer vision},
  pages={15013--15022},
  year={2021}
}

@article{personreid,
  title={Deep learning for person re-identification: A survey and outlook},
  author={Ye, Mang and Shen, Jianbing and Lin, Gaojie and Xiang, Tao and Shao, Ling and Hoi, Steven CH},
  journal={IEEE transactions on pattern analysis and machine intelligence},
  volume={44},
  number={6},
  pages={2872--2893},
  year={2021},
  publisher={IEEE}
}

@article{chatreid,
  title={Chatreid: Open-ended interactive person retrieval via hierarchical progressive tuning for vision language models},
  author={Niu, Ke and Yu, Haiyang and Zhao, Mengyang and Fu, Teng and Yi, Siyang and Lu, Wei and Li, Bin and Qian, Xuelin and Xue, Xiangyang},
  journal={arXiv preprint arXiv:2502.19958},
  year={2025}
}

@article{idavlm,
  title={Ida-vlm: Towards movie understanding via id-aware large vision-language model},
  author={Ji, Yatai and Zhang, Shilong and Wu, Jie and Sun, Peize and Chen, Weifeng and Xiao, Xuefeng and Yang, Sidi and Yang, Yujiu and Luo, Ping},
  journal={arXiv preprint arXiv:2407.07577},
  year={2024}
}

@inproceedings{ilr-met,
  title={The met dataset: Instance-level recognition for artworks},
  author={Ypsilantis, Nikolaos-Antonios and Garcia, Noa and Han, Guangxing and Ibrahimi, Sarah and Van Noord, Nanne and Tolias, Giorgos},
  booktitle={Thirty-fifth conference on neural information processing systems datasets and benchmarks track (Round 2)},
  year={2021}
}

@inproceedings{ilr-landmark,
  title={Revisiting oxford and paris: Large-scale image retrieval benchmarking},
  author={Radenovi{\'c}, Filip and Iscen, Ahmet and Tolias, Giorgos and Avrithis, Yannis and Chum, Ond{\v{r}}ej},
  booktitle={Proceedings of the IEEE conference on computer vision and pattern recognition},
  pages={5706--5715},
  year={2018}
}

@inproceedings{sop,
  title={Deep metric learning via lifted structured feature embedding},
  author={Oh Song, Hyun and Xiang, Yu and Jegelka, Stefanie and Savarese, Silvio},
  booktitle={Proceedings of the IEEE conference on computer vision and pattern recognition},
  pages={4004--4012},
  year={2016}
}

@inproceedings{ilr-google-landmark,
  title={Google landmarks dataset v2-a large-scale benchmark for instance-level recognition and retrieval},
  author={Weyand, Tobias and Araujo, Andre and Cao, Bingyi and Sim, Jack},
  booktitle={Proceedings of the IEEE/CVF conference on computer vision and pattern recognition},
  pages={2575--2584},
  year={2020}
}

@inproceedings{petface,
  title={Petface: A large-scale dataset and benchmark for animal identification},
  author={Shinoda, Risa and Shiohara, Kaede},
  booktitle={European Conference on Computer Vision},
  pages={19--36},
  year={2024},
  organization={Springer}
}

@inproceedings{vggface,
  title={Vggface2: A dataset for recognising faces across pose and age},
  author={Cao, Qiong and Shen, Li and Xie, Weidi and Parkhi, Omkar M and Zisserman, Andrew},
  booktitle={2018 13th IEEE international conference on automatic face \& gesture recognition (FG 2018)},
  pages={67--74},
  year={2018},
  organization={IEEE}
}

@article{deepseekvl,
  title={Deepseek-vl: towards real-world vision-language understanding},
  author={Lu, Haoyu and Liu, Wen and Zhang, Bo and Wang, Bingxuan and Dong, Kai and Liu, Bo and Sun, Jingxiang and Ren, Tongzheng and Li, Zhuoshu and Yang, Hao and others},
  journal={arXiv preprint arXiv:2403.05525},
  year={2024}
}

@inproceedings{clip,
  title={Learning transferable visual models from natural language supervision},
  author={Radford, Alec and Kim, Jong Wook and Hallacy, Chris and Ramesh, Aditya and Goh, Gabriel and Agarwal, Sandhini and Sastry, Girish and Askell, Amanda and Mishkin, Pamela and Clark, Jack and others},
  booktitle={International conference on machine learning},
  pages={8748--8763},
  year={2021},
  organization={PmLR}
}

@inproceedings{siglip,
  title={Sigmoid loss for language image pre-training},
  author={Zhai, Xiaohua and Mustafa, Basil and Kolesnikov, Alexander and Beyer, Lucas},
  booktitle={Proceedings of the IEEE/CVF international conference on computer vision},
  pages={11975--11986},
  year={2023}
}

@article{siglip2,
  title={Siglip 2: Multilingual vision-language encoders with improved semantic understanding, localization, and dense features},
  author={Tschannen, Michael and Gritsenko, Alexey and Wang, Xiao and Naeem, Muhammad Ferjad and Alabdulmohsin, Ibrahim and Parthasarathy, Nikhil and Evans, Talfan and Beyer, Lucas and Xia, Ye and Mustafa, Basil and others},
  journal={arXiv preprint arXiv:2502.14786},
  year={2025}
}

@inproceedings{dino,
  title={Emerging properties in self-supervised vision transformers},
  author={Caron, Mathilde and Touvron, Hugo and Misra, Ishan and J{\'e}gou, Herv{\'e} and Mairal, Julien and Bojanowski, Piotr and Joulin, Armand},
  booktitle={Proceedings of the IEEE/CVF international conference on computer vision},
  pages={9650--9660},
  year={2021}
}

@article{dinov2,
  title={Dinov2: Learning robust visual features without supervision},
  author={Oquab, Maxime and Darcet, Timoth{\'e}e and Moutakanni, Th{\'e}o and Vo, Huy and Szafraniec, Marc and Khalidov, Vasil and Fernandez, Pierre and Haziza, Daniel and Massa, Francisco and El-Nouby, Alaaeldin and others},
  journal={arXiv preprint arXiv:2304.07193},
  year={2023}
}

@article{prcc,
  title={Person Re-identification by Contour Sketch under Moderate Clothing Change},
  author={Yang, Qize and Wu, Ancong and Zheng, Wei-Shi},
  journal={IEEE Transactions on Pattern Analysis and Machine Intelligence (TPAMI)},
  year={2020},
  volume={42},
  number={10},
  pages={2577-2591},
  doi={10.1109/TPAMI.2019.2960509}
}

@inproceedings{lpw,
  author = {Guanglu Song and Biao Leng and Yu Liu and Congrui Hetang and Shaofan Cai},
  title={Region-based Quality Estimation Network for Large-Scale Person Re-identiﬁcation},
  booktitle={AAAI Conference on Artificial Intelligence (AAAI)},
  year = {2018}
}

@inproceedings{cuhk,
  title={DeepReID: Deep Filter Pairing Neural Network for Person Re-identification},
  author={Li, Wei and Zhao, Rui and Xiao, Tong and Wang, Xiaogang},
  booktitle={IEEE Conference on Computer Vision and Pattern Recognition (CVPR)},
  year={2014}
}

@inproceedings{market,
  title={Scalable Person Re-identification: A Benchmark},
  author={Zheng, Liang and Shen, Liyue and Tian, Lu and Wang, Shengjin and Wang, Jingdong and Tian, Qi},
  booktitle={IEEE International Conference on Computer Vision (ICCV)},
  year={2015},
  pages={1116-1124}
}

@inproceedings{rstp,
  title={RSTPReid: A Large-Scale Dataset for Text-based Person Re-identification},
  author={Zhu, Yanlin and Wu, Xinyu and Su, Mingjia and Wang, Zhiming and Zhang, Zhicheng and Yin, Guojun},
  booktitle={International Conference on Pattern Recognition (ICPR)},
  year={2021}
}

@article{attention,
  title={Attention is all you need},
  author={Vaswani, Ashish and Shazeer, Noam and Parmar, Niki and Uszkoreit, Jakob and Jones, Llion and Gomez, Aidan N and Kaiser, {\L}ukasz and Polosukhin, Illia},
  journal={Advances in neural information processing systems},
  volume={30},
  year={2017}
}

@article{plip,
  title={Plip: Language-image pre-training for person representation learning},
  author={Zuo, Jialong and Hong, Jiahao and Zhang, Feng and Yu, Changqian and Zhou, Hanyu and Gao, Changxin and Sang, Nong and Wang, Jingdong},
  journal={Advances in Neural Information Processing Systems},
  volume={37},
  pages={45666--45702},
  year={2024}
}

@inproceedings{arcface,
  title={Arcface: Additive angular margin loss for deep face recognition},
  author={Deng, Jiankang and Guo, Jia and Xue, Niannan and Zafeiriou, Stefanos},
  booktitle={Proceedings of the IEEE/CVF conference on computer vision and pattern recognition},
  pages={4690--4699},
  year={2019}
}

@inproceedings{clipscore,
  title={Clipscore: A reference-free evaluation metric for image captioning},
  author={Hessel, Jack and Holtzman, Ari and Forbes, Maxwell and Le Bras, Ronan and Choi, Yejin},
  booktitle={Proceedings of the 2021 conference on empirical methods in natural language processing},
  pages={7514--7528},
  year={2021}
}
}

\clearpage
\setcounter{page}{1}
\maketitlesupplementary

\begin{strip}
    \centering
    \includegraphics[width=1\textwidth]{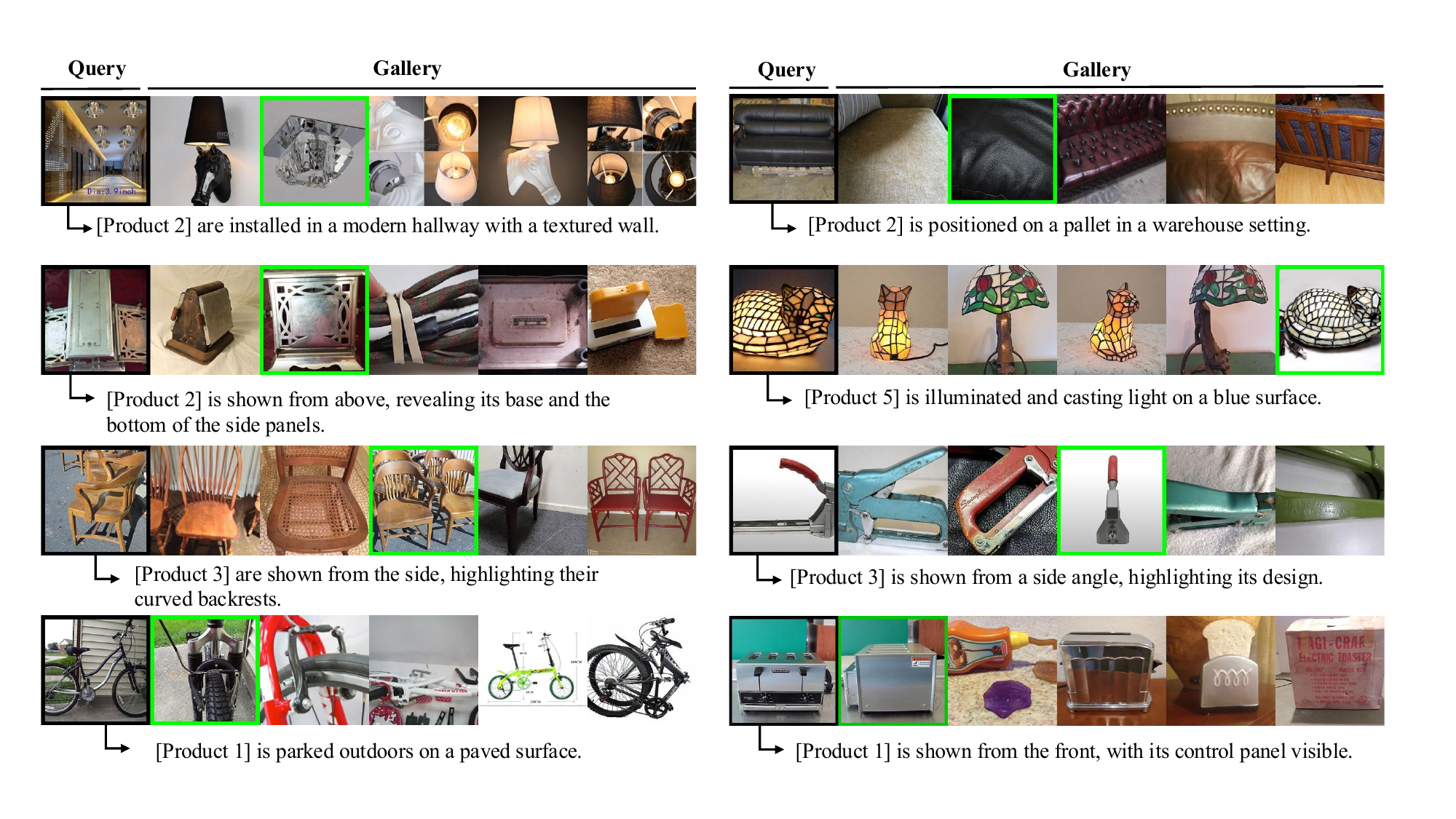}
    \vspace{-2em}
    \captionof{figure}
    {\textbf{Additional examples of our instance-level recognition dataset along with corresponding model outputs. } IIR-VLM is consistently capable of correctly identifying the matching item in the gallery (green frame) and referring to the item in caption generation.}

    \label{fig:extension}
\end{strip}

\section{Appendix}

In the appendix, we provide additional visualizations of our instance-level recognition dataset alongside the corresponding IIR-VLM outputs. We include expanded examples, illustrate the varying difficulty levels of the dataset, and present an analysis of failure cases.

\subsection{Additional visualizations}

In~\cref{fig:extension}, we provide an extension of~\cref{fig:captions} featuring additional examples from the Stanford Online Product~\cite{sop} dataset, which makes up the object category in our benchmark. Our IIR-VLM performs consistently well on this diverse category, achieving an accuracy of 91.2\% when boosted by the auxiliary expert encoder. All examples showcase the correct selection of the matching instance. While some cases are straightforward, many require acute attention to detail and the ability to associate different local features or disparate viewpoints of the objects. Throughout these examples, the generated captions consistently refer to the correct object and provide accurate and useful descriptions of the context.

\subsection{Visualizing similarity filtering}

\begin{figure*}[t]
    \centering
    \includegraphics[width=1\textwidth]{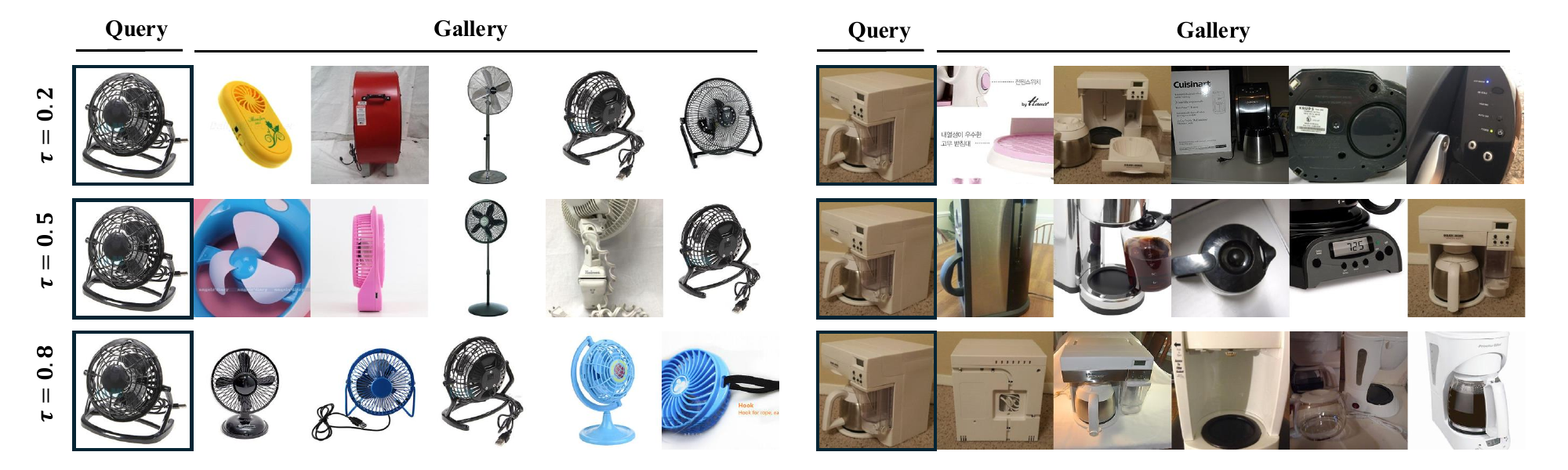}
    \caption{\textbf{Visualizing the effect of the similarity threshold $\tau$}, which controls the visual similarity between gallery images and the query image. As the threshold increases, the gallery is populated with candidates more similar to the query, thereby making the benchmark more challenging. }
    \label{fig:difficulty}
\end{figure*}

In~\cref{exp:ablation}, we demonstrate that the expert encoder provides an additional advantage as the benchmark difficulty increases. Here in~\cref{fig:difficulty}, we provide visual examples of these difficulty levels, which are controlled by the similarity threshold $\tau$. In both examples, we fix the query image and select gallery images that satisfy specific visual similarity thresholds ($\tau=0.2, 0.5, 0.8$). 

The task of identifying matching instances is fairly straightforward with $\tau=0.2$, given that the visual appearances of selected images are often drastically different from one another. However, as $\tau$ increases, features such as structure and color increasingly resemble the query image. Successfully addressing these difficult cases necessitates the specialized knowledge of category-specific expert ILR models.

\begin{figure*}[t]
    \centering
    \includegraphics[width=1\textwidth]{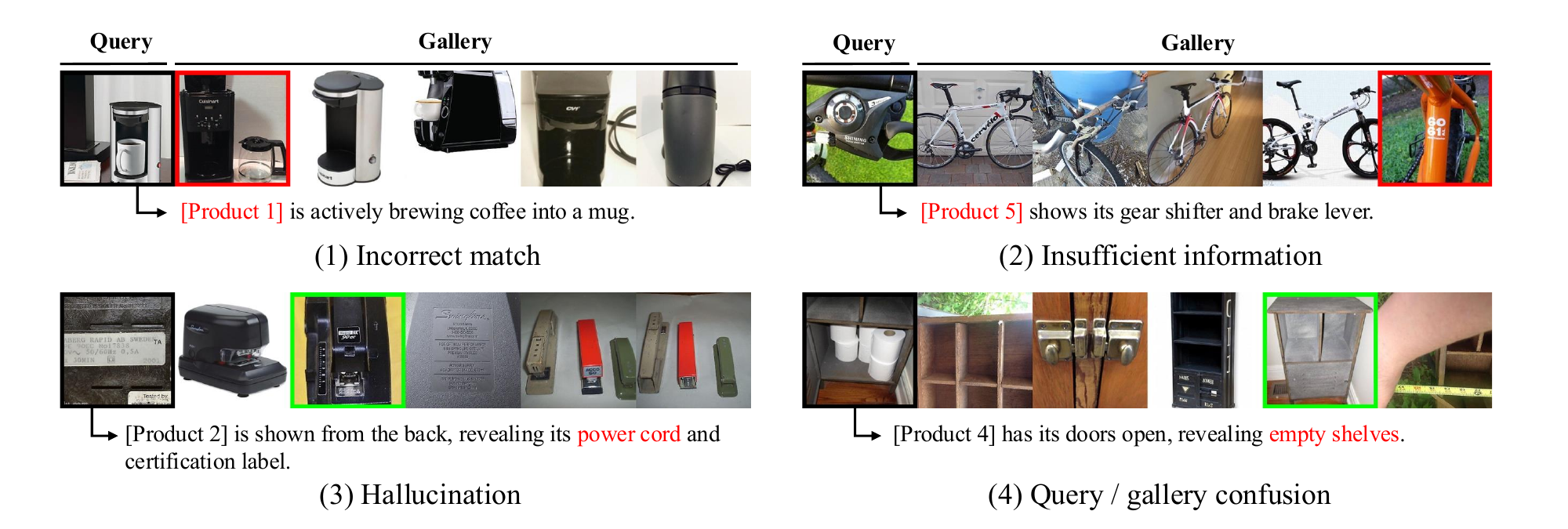}
    \caption{\textbf{Failure cases. } We identify different types of failure cases that arise in our instance-level recognition and understanding tasks. Incorrect selections and inaccurate descriptions are marked in red.}
    \label{fig:failures}
\end{figure*}

\subsection{Failure cases}

We present some failure cases of our method in~\cref{fig:failures}. These failures are classified based on the stage of our pipeline, separating failures to identify the correct instance from failures to  provide accurate descriptions.

\textbf{Instance matching failures. }
As indicated by an error rate of approximately 10\%, models still make mistakes in identifying matching instances. Even with an expert encoder, extreme visual similarity remains a bottleneck. For instance, in Case (1), a non-matching gallery image shares a highly similar layout with the query, leading to confusion. Future work needs to focus on even stronger fine-grained visual understanding to disentangle variables like scene layout from object identity.

We also observe benchmark-level challenges where insufficient visual overlap is provided for the ILR problem. In Case (2), the gallery provides mostly global views of bicycles while the query features an extreme close-up, which could match any gallery instance. We view these cases as an inadequate data curation issue, and not a failure of the model's visual understanding capabilities.

\textbf{Image understanding failures. } We also observe cases where the model successfully identifies the correct instance but fails to generate an accurate description. In Case (3), the model exhibits typical hallucination, describing a "power cord" that does not exist in the image, possibly driven by language priors rather than visual evidence. Additionally, models tend to confuse different images in the context. Among many examples, the model in Case (4) generates a caption that resembles the matched gallery image more than the query image. These failures highlight that while our method improves recognition, it still inherits the general limitations of current VLM backbones. Evidently, the success in instance-level understanding requires parallel progress in fundamental visual understanding capabilities. 

\end{document}